\documentclass{article}

\usepackage{microtype}
\usepackage{graphicx}
\usepackage{subfigure}
\usepackage{booktabs} 
\usepackage{times}
\usepackage{epsfig}
\usepackage{amsmath}
\usepackage{amssymb}
\usepackage{bm}
\usepackage{amsthm}
\usepackage{multirow}
\usepackage[super]{nth}

\usepackage{amssymb}%
\usepackage{mathrsfs}%
\usepackage{nicefrac} 
\usepackage{hyperref}

\usepackage[accepted]{icml2019}

\newtheorem{theorem}{Theorem}
\newtheorem{corollary}{Corollary}

\newtheorem{theoremappendix}{Theorem*}

\DeclareMathOperator{\sgn}{sign}
\DeclareMathOperator{\softmax}{\mathbb{S}}
\DeclareMathOperator{\R}{\mathbb{R}}
\DeclareMathOperator{\H_en}{\mathcal{H}}
\DeclareMathOperator{\F_en}{\mathcal{F}}
\DeclareMathOperator{\Loss}{\mathcal{L}}
\DeclareMathOperator{\CE}{\text{CE}}
\DeclareMathOperator{\ECE}{\text{ECE}}
\DeclareMathOperator{\JSD}{\text{JSD}}
\DeclareMathOperator{\ADP}{\text{ADP}}

\DeclareMathOperator{\clip}{clip}
\DeclareMathOperator{\stt}{s.t.}
\DeclareMathOperator{\vol}{\text{Vol}}
\DeclareMathOperator{\ED}{\mathbb{ED}}

\hypersetup{
	colorlinks=true,
	urlcolor=magenta,
}
\icmltitlerunning{Improving Adversarial Robustness via Promoting Ensemble Diversity}

\begin{document}

\twocolumn[
\icmltitle{Improving Adversarial Robustness via Promoting Ensemble Diversity}




\begin{icmlauthorlist}
\icmlauthor{Tianyu Pang}{to}
\icmlauthor{Kun Xu}{to}
\icmlauthor{Chao Du}{to}
\icmlauthor{Ning Chen}{to}
\icmlauthor{Jun Zhu}{to}
\end{icmlauthorlist}

\icmlaffiliation{to}{Department of Computer Science and Technology, Institute for AI, BNRist Center, THBI Lab, Tsinghua-Fuzhou Institute for Data Technology, Tsinghua University, Beijing, China}

\icmlcorrespondingauthor{Tianyu Pang}{pty17@mails.tsinghua.edu.cn}
\icmlcorrespondingauthor{Ning Chen}{ningchen@tsinghua.edu.cn}
\icmlcorrespondingauthor{Jun Zhu}{dcszj@tsinghua.edu.cn}

\icmlkeywords{Machine Learning, ICML}

\vskip 0.3in
]



\printAffiliationsAndNotice{}  

\begin{abstract}
Though deep neural networks have achieved significant progress on various tasks, often enhanced by model ensemble, existing high-performance models can be vulnerable to adversarial attacks. Many efforts have been devoted to enhancing the robustness of individual networks and then constructing a straightforward ensemble, e.g., by directly averaging the outputs, which ignores the interaction among networks. This paper presents a new method that explores the interaction among individual networks to improve robustness for ensemble models. Technically, we define a new notion of ensemble diversity in the adversarial setting as the diversity among \emph{non-maximal} predictions of individual members, and present an adaptive diversity promoting (ADP) regularizer to encourage the diversity, which leads to globally better robustness for the ensemble by making adversarial examples difficult to transfer among individual members. Our method is computationally efficient and compatible with the defense methods acting on individual networks. Empirical results on various datasets verify that our method can improve adversarial robustness while maintaining state-of-the-art accuracy on normal examples.

\end{abstract}

\vspace{-0.45cm}
\section{Introduction}
\vspace{-0.05cm}
Deep neural networks (DNNs) have recently achieved significant progress in many tasks, such as image classification~\cite{russakovsky2015imagenet} and speech recognition~\cite{graves2014towards}. However, they are not problemless. Recent work has shown that high-performance classifiers are vulnerable to adversarial attacks~\cite{goodfellow6572explaining,papernot2016limitations,carlini2017towards,athalye2018obfuscated}, where images with human imperceptible perturbations (i.e., adversarial examples) can be generated to fool the classifiers. To improve the adversarial robustness of classifiers, various kinds of defenses have been proposed~\cite{kurakin2016atscale,kurakin2018ensemble,kannan2018adversarial}, and one practically effective strategy is to construct ensembles of the enhanced networks to obtain stronger defenses~\cite{liao2017defense,kurakin2018competation}. 

However, most of these defense strategies focus on enhancing an individual network, while ignoring the potentially very informative interactions among multiple models. As widely observed, adversarial examples can have strong transferability among models~\cite{papernot2016practical,kurakin2018competation}. When training an ensemble model, the ignorance of interaction among individual members may lead them to return similar predictions or feature representations~\cite{dauphin2014identifying,li2016convergent}. Since it is easier for adversarial examples to transfer among similar models, it is probable for the adversarial examples crafted for one individual member to also fool other members, or even fool the whole ensemble model (See experiments).



\begin{figure*}[t]
\begin{center}
\vspace{-0.2cm}
\includegraphics[width=2.00\columnwidth]{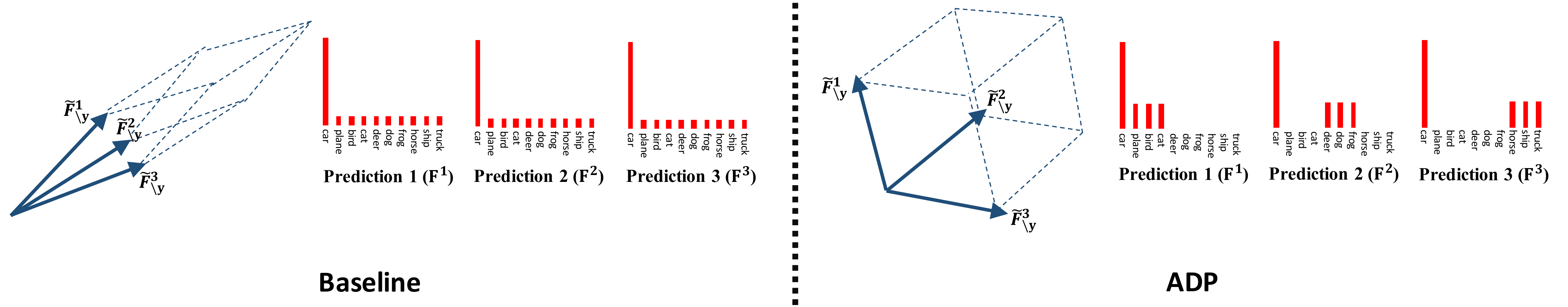}
\vspace{-0.4cm}
\caption{Illustration of the ensemble diversity. \textbf{Baseline:} Individually training each member of the ensemble. \textbf{ADP:} Simultaneously training all the members of the ensemble with the ADP regularizer. The left part of each panel is the normalized non-maximal predictions.}
\label{fig:2}
\end{center}
\vspace{-0.4cm}
\end{figure*}

Previous work has shown that for a single network, promoting the diversity among learned features of different classes can improve adversarial robustness~\cite{pang2017towards,pang2018max}. In this paper, we propose to improve adversarial robustness from a new perspective by promoting the diversity among the predictions returned by different members of an ensemble, called {\it ensemble diversity}. Our approach is orthogonal to other defenses acting on a single network and thus is compatible with most of the previous defense methods. 

Technically, we first develop a new notion of ensemble diversity in the adversarial setting, which is significantly different from the previous attempts in the standard setting of ensembling weak classifiers in a non-adversarial environment. Namely, previous work has defined ensemble diversity by considering the prediction errors of individual classifiers in the ensemble~\cite{liu1999ensemble,liu1999simultaneous,dietterich2000ensemble,islam2003constructive}. However, this definition is inappropriate for our case since most DNNs can achieve high accuracy on a training set~\cite{goodfellow2016deep} and promoting diversity of prediction errors will largely sacrifice accuracy. Instead, we define the diversity on the \emph{non-maximal} predictions of each network (detailed in Eq.~(\ref{equation:12})). Geometrically, the diversity equals to the squared volume spanned by the normalized non-maximal predictions, as illustrated in Fig.~\ref{fig:2}. Note that promoting the diversity still allows the maximal prediction of each network to be consistent with the true label, and thus will not affect ensemble accuracy. Besides,
since the non-maximal predictions correspond to all potentially wrong labels returned for the adversarial examples, a high diversity or inconsistency on the non-maximal predictions of individual networks can lower down the transferability of adversarial examples among them, and further lead to better robustness of the ensemble.

Under our definition of ensemble diversity, we propose to promote the diversity using an {\it adaptive diversity promoting} (ADP) regularizer. The ADP regularizer consists of two terms---a logarithm of ensemble diversity (LED) term and an ensemble entropy term. The ensemble entropy term contains the Shannon entropy of ensemble predictions. In Section~\ref{Theoretical}, we prove that the ensemble entropy term is necessary such that the LED term can properly encourage the diversity. The ADP regularizer encourages the non-maximal predictions of each member in the ensemble to be mutually orthogonal, while keeping the maximal prediction be consistent with the true label, as shown in Fig.~\ref{fig:2} and theoretically proved in Section~\ref{Theoretical}. 
In the training procedure, we use the ADP regularizer as the penalty term and train all the networks in the ensemble simultaneously and interactively on the same dataset. This is called the ADP training procedure.



In experiments, we test our method under several widely studied adversarial attacks introduced in Section~\ref{attacks} on MNIST~\cite{lecun1998gradient}, CIFAR-10, and CIFAR-100~\cite{Krizhevsky2012}. The results show that our method can significantly improve adversarial robustness while maintaining state-of-the-art accuracy on normal examples. Although when performing back-propagation~\cite{rumelhart1986learning} in the ADP training procedure, we need to separately perform the operations $\det(G)$ and $G^{-1}$ with the computational complexity of $\mathcal{O}(K^3)$, where $K$ is the number of members in the ensemble, $K$ usually grows much slower with the scale of problem~\cite{russakovsky2015imagenet}. This enables our method to scale to most of the classification tasks and avoid the computational obstacle caused by these matrix operations as encountered in previous work~\cite{kulesza2012determinantal,kwok2012priors,mariet2016diversity}.


\vspace{-0.cm}
\section{Preliminary Knowledge}
\vspace{-0.cm}

In this section, we first define the notations, then briefly introduce the adversarial attacks that we test in experiments.

\subsection{Notations}
We denote a deep neural network (DNN) classifier as $F(x,\theta)$, where $x$ is the input variable, and $\theta$ denotes all the trainable parameters. Let $L$ be the number of classes, the output vector $F(x,\theta)\in\R^L$ represents a probability distribution (Hereafter we will omit $\theta$ without ambiguity). When training a classifier, one common objective function is the cross-entropy (CE) loss defined as
\[\Loss_{\text{CE}}(x, y)=-1_y^\top  \log{F(x)}=-\log{F(x)_y}\text{,}\]
for an input-label pair $(x,y)$. Here, $1_y$ is the one-hot encoding of $y$ and the logarithm of a vector is defined as taking logarithm of each element. Given a probability vector $F$, the Shannon entropy is defined as $\H_en(F)=-\sum_iF_i\log(F_i)$.

\subsection{The Adversarial Setting}
\label{attacks}
Recent work shows that high-performance classifiers could still be fooled by adversarial examples~\cite{kurakin2018competation,athalye2018obfuscated}. A generated adversarial example $x^*$ is usually indistinguishable from its normal counterpart $x$ by a human observer. Here we introduce some of the most commonly used attack methods as below:

\textbf{Fast Gradient Sign Method (FGSM)}~\cite{goodfellow6572explaining} crafts the adversarial example $x^*$ as
$x^*=x+\epsilon\cdot\sgn(\nabla_{x}\mathcal{L}(x, y))$, where $\mathcal{L}(x, y)$ is the adversarial loss.

\textbf{Basic Iterative Method (BIM)}~\cite{kurakin2016adversarial} iteratively crafts an adversarial example as
$x^*_i=\clip_{x,\epsilon}(x^*_{i-1}+\frac{\epsilon}{r}\cdot\sgn(\nabla_{x^*_{i-1}}\mathcal{L}(x^*_{i-1}, y)))$. Here $x^*_0=x$, $r$ is the number of iteration steps and $\clip_{x,\epsilon}(\cdot)$ is the clipping function.

\textbf{Projected Gradient Descent (PGD)}~\cite{madry2017towards} has the same generation process as BIM, except for $x_{0}^*$ is a randomly perturbed image in the neighborhood $\mathring{U}(x,\epsilon)$. 

\textbf{Momentum Iterative Method (MIM)}~\cite{dong2017boosting} won the NeurIPS 2017 Adversarial Competition~\cite{kurakin2018competation}, which is a variant of BIM. In each iteration, MIM updates the gradient $g_i$ with momentum and crafts the adversarial example as $x_{i}^*=\clip_{x,\epsilon}(x^*_{i-1}+\frac{\epsilon}{r}\cdot\sgn(g_i))$

\textbf{Jacobian-based Saliency Map Attack (JSMA)}~\cite{papernot2016limitations} perturbs one feature $x_i$ by $\epsilon$ in each iteration step that maximizes the saliency map
\vspace{-0.2cm}
\[S(x,y)[i]=\begin{cases}
0\text{, if } \frac{\partial F(x)_y}{\partial x_i}<0\text{ or }\sum_{j\neq y}\frac{\partial F(x)_j}{\partial x_i}>0\text{,}\\
(\frac{\partial F(x)_y}{\partial x_i})\left| \sum_{j\neq y}\frac{\partial F(x)_j}{\partial x_i} \right|\text{,otherwise.}
\end{cases}\]
The adversarial noise crafted by JSMA is often sparser, i.e., JSMA perturbs fewer pixels compared to other attacks.

\textbf{Carlini \& Wagner (C\&W)}~\cite{carlini2017towards} define $x^*(\omega)=\frac{1}{2}(\tanh({\omega})+1)$ in terms of $\omega$, and minimize the objective function
$\min_\omega \|x^*(\omega)-x\|_2^2+c\cdot f(x^*(\omega))$,
where $c$ is a constant chosen by a modified binary search. Let $\softmax_{pre}(x)$ be the input vector of the softmax function in a classifier, then $f(\cdot)$ is the objective function defined as
\vspace{-0.1cm}
\[f(x)=\max(\max\{\softmax_{pre}(x)_i:i\neq y \}-\softmax_{pre}(x)_i, -\kappa)\text{,}\]
where $\kappa$ controls the confidence on adversarial examples.

\textbf{Elastic-net Attack (EAD)}~\cite{chen2017ead} is a variant of the C\&W method, where the objective function becomes $\min_\omega \|x^*(\omega)-x\|_2^2+\beta\|x^*(\omega)-x\|_1+c\cdot f(x^*(\omega))$. EAD includes C\&W as a special case when $\beta=0$.


\section{Methodology}
In this section, we first introduce the training strategies for ensemble models. Then we propose the adaptive diversity promoting (ADP) training method, and provide theoretical analyses on the optimal solutions of the ADP method.
 

\subsection{Training Strategies for Ensemble Models}
\label{train_strategies}
\vspace{-0.1cm}

In real-world problems, ensembling several individual classifiers is usually effective to improve generalization performance~\cite{russakovsky2015imagenet}. For clarity, we denote the $k$-th member in the ensemble as $F^{k}\in\R^{L}$, which represents the prediction of classifier $k$ over $L$ categories. Then a common construction of the ensemble prediction $\F_en$ is a simple average of all individual predictions as
\vspace{-0.15cm}
\begin{equation}
    \F_en=\frac{1}{K}\sum_{k\in[K]}F^{k}\text{,}
    \vspace{-0.15cm}
\end{equation}
where $[K]:=\{1, \cdots, K\}$. There are three major strategies for training ensembles, i.e., independent training, simultaneous training, and sequential training~\cite{islam2003constructive}.

In independent training, each classifier $F^k$ is trained separately, usually with the corresponding CE loss $\Loss_{\CE}^{k}(x,y):=-\log F^{k}(x)_y$. Because of its simplicity, most of the deep learning systems apply this strategy~\cite{russakovsky2015imagenet,liao2017defense}. However, one major disadvantage of independent training is that it ignores the interaction among the members in the ensemble. This ignorance may lead to similar predictions or shared feature representations among these members~\cite{dauphin2014identifying,li2016convergent}, which would not be useful to improve the performance of the ensemble~\cite{krogh1995neural}, and result in less improvement on robustness as shown in our experiments.

In simultaneous training, all the classifiers are trained on the same mini-batch of data in each training iteration. A naive objective function for simultaneous training is the ensemble cross-entropy (ECE) loss defined as
\vspace{-0.1cm}
\begin{equation}
    \Loss_{\ECE}=\sum_{k\in[K]}\Loss_{\CE}^{k}\text{,}
    \vspace{-0.1cm}
\end{equation}
which is a simple sum of all individual cross-entropy losses. Since the parameters for different networks are typically independent, simultaneously training all $K$ networks with the ECE loss is equivalent to independently training each classifier with the corresponding CE loss. The learning rate for each network could be different. If some networks converge faster, then we can freeze the parameters of those converged networks and continue training others. Simultaneous training scheme allows us to incorporate the interaction among different members in the ensemble. Indeed, for large-scale tasks, simultaneous training may require more GPU memory or workers compared to independent training. However, it does not require extra computation. When the computing resource is limited, sequential training can be applied instead, which trains one network after another in each iteration.

In our method, we apply the simultaneous training scheme. Namely, we combine a novel regularizer (introduced in Eq.~(\ref{ADP})) with the ECE loss term as the objective function for training. By doing this, the members in the ensemble can separately keep high accuracy since the ECE loss term, while interacting with each other via our regularizer.

\subsection{Adaptive Diversity Promoting Training}
\vspace{-0.1cm}

Previous work has shown that for a single network, promoting the diversity among learned features of different classes can improve adversarial robustness~\cite{pang2017towards,pang2018max}. 
In this paper, we promote diversity from another level, i.e., the diversity among learned predictions of different members in an ensemble, to improve robustness. This kind of diversity is called \emph{ensemble diversity}. The previous analysis states that there is no strict definition of what is intuitively perceived as ensemble diversity~\cite{kuncheva2003measures} and most of the previous work defines the ensemble diversity concerning the prediction errors of individual classifiers in the ensemble~\cite{liu1999ensemble,liu1999simultaneous,dietterich2000ensemble,islam2003constructive}. This definition partly bases on the assumption that each member in the ensemble is a weak classifier~\cite{friedman2001elements}. However, this assumption hardly holds for the ensemble of DNNs nowadays, thus encouraging the prediction errors of DNNs to be diverse is inappropriate and will largely sacrifice the accuracy of them.




In order to maintain the accuracy of each member in the ensemble, the maximal prediction returned by each member should be consistent with the true label. Therefore, in our method, we define the ensemble diversity with respect to the \emph{non-maximal} predictions. For a single input pair $(x,y)$ in the training set, the maximal prediction should be consistent with the true label $y$, and the non-maximal predictions correspond to other labels except for $y$. Inspired by the theory of the determinant point process (DPP)~\cite{kulesza2012determinantal}, we define the ensemble diversity as
\vspace{-0.cm}
\begin{equation}
    \ED=\det(\tilde{M}_{\backslash y}^{\top} \tilde{M}_{\backslash y})\text{.}
    \label{equation:12}
\end{equation}
Here $\tilde{M}_{\backslash y}=(\tilde{F}^{1}_{\backslash y},\cdots,\tilde{F}^{K}_{\backslash y})\in\R^{(L-1)\times K}$, each column vector $\tilde{F}^{k}_{\backslash y}\in\R^{L-1}$ is obtained by normalizing $F^{k}_{\backslash y}$ under $L_2$-norm, where $F^{k}_{\backslash y}$ is the order preserving prediction of the $k$-th classifier on $x$ without the $y$-th element. According to the matrix theory~\cite{bernstein2005matrix}, we have:
\vspace{-0.cm}
\begin{equation}
    \det(\tilde{M}_{\backslash y}^{\top} \tilde{M}_{\backslash y})=\vol^{2}(\{\tilde{F}^{k}_{\backslash y}\}_{k\in[K]})\text{,}
\end{equation}
where $\vol(\cdot)$ denotes the volume spanned by the vectors of the input set. This provides an elegant geometric interpretation for the ensemble diversity, as intuitively shown in Fig.~\ref{fig:2}. Since $\tilde{F}^{k}_{\backslash y}$ is normalized, i.e., $\|\tilde{F}^{k}_{\backslash y}\|_2=1$, the ensemble diversity $\ED$ achieves its maximal value $1$ if and only if the column vectors of $\tilde{M}_{\backslash y}$ are mutually orthogonal.

Note that most of the previous definitions of diversity, e.g., those defined on prediction errors, include the predictions $F_{y}^{k}$ on the true label $y$~\cite{liu1999ensemble,liu1999simultaneous,islam2003constructive,kuncheva2003measures}. Since DNNs are no longer weak classifiers in the most cases~\cite{goodfellow2016deep}, encouraging $F_{y}^{k}$ to be diverse may make the loss on accuracy outweighs the gain for DNNs, or render the convergence point of the training process uncontrollable. In contrast, we define the diversity among the normalized non-maximal predictions $\tilde{F}^{k}_{\backslash y}$, which allows the maximal prediction of each network to be consistent with the true label, and thus will not affect ensemble accuracy. What's more, promoting this diversity could improve robustness. Because, in the adversarial setting, when the maximal prediction corresponds to the true label $y$, the non-maximal predictions correspond to the labels $[L]\backslash\{y\}$, which include all potentially wrong labels returned for the adversarial examples. Thus a high diversity or inconsistency on the non-maximal predictions $F^{k}_{\backslash y}$ can lower down the transferability of adversarial examples among the networks, and further lead to better robustness of the ensemble.

\begin{algorithm}[t]
\caption{The ADP training procedure}
\label{algo:2}
\begin{algorithmic}
\STATE {\bfseries Input:} The $K$ individual models $\{F^{k}(x,\theta^{k})\}_{k\in[K]}$; the training dataset $\mathcal{D}=\{(x_i, y_i)\}_{i\in [N]}$.\\
\STATE {\bfseries Initialization:} Initialize each $\theta^k$ as $\theta^k_0$, the training step counters as $c_k=0$, $\varepsilon_k$ be the learning rate variables, $I=[K]$ be the indicator set, where $k\in[K]$.
\WHILE{$I\neq\varnothing$}
\STATE Calculate the objective on a mini-batch of data $\mathcal{D}_m$
\vspace{-0.3cm}
\[\mathcal{L}_{\text{ADP}}^m=\frac{1}{\left|\mathcal{D}_m\right|} \sum_{(x_i, y_i) \in \mathcal{D}_m}\left[\mathcal{L}_{\text{ECE}}-\ADP_{\alpha,\beta}\right](x_i,y_i)\text{.}\]
\vspace{-0.45cm}
\FOR{$k^{'}$ in $I$}
\STATE Update $\theta^{k^{'}}_{c+1} \leftarrow\theta^{k^{'}}_{c}-\varepsilon_{k^{'}}\nabla_{\theta^{k^{'}}}\mathcal{L}_{\text{ADP}}^m\big|_{\{\theta^{k}_{c_k}\}_{k\in[K]}}$.
\STATE Update $c_{k^{'}}\leftarrow c_{k^{'}}+1$, where $c=c_{k^{'}}$.
\STATE \textbf{if} $\theta^{k^{'}}$ converges \textbf{then} Update $I=I\backslash\{k^{'}\}$.
\ENDFOR 
\ENDWHILE
\STATE {\bfseries Return:} The parameters $\theta^{k}=\theta^{k}_{c_k}$, where $k\in[K]$.
\end{algorithmic}
\end{algorithm}

To promote ensemble diversity, we propose the \textbf{adaptive diversity promoting (ADP)} regularizer as
\begin{equation}
\label{ADP}
\ADP_{\alpha,\beta}(x,y)=\alpha\cdot\H_en({\F_en})+\beta\cdot\log\left(\ED\right)
\end{equation}
for a single input pair $(x,y)$, where $\alpha,\beta \geq 0$ are two hyperparameters. Note that the ADP regularizer consists of two parts. The first part is the ensemble Shannon entropy $\H_en({\F_en})$, 
and the second part is the logarithm of the ensemble diversity (LED), which we call as the LED part. More technical details about the effects of these two parts and why the ensemble entropy part is necessary will be deferred to Section~\ref{Theoretical}. The training procedure with the ADP regularizer is called the ADP training method, as described in Algorithm~\ref{algo:2}. Further analysis on other potential definition of ensemble diversity is provided in Appendix B.1.

When performing back-propagation~\cite{rumelhart1986learning} in the ADP training procedure, we need to separately perform two matrix operations $\det(G)$ and $G^{-1}$. Without loss of generality, let $G\in\R^{m\times m}$, where $m$ denote the dimension of the matrix, these two operations both have computational complexities of $\mathcal{O}(m^3)$ under common algorithms~\cite{kulesza2012determinantal}, which may cause computational obstacle if $m$ scales comparably with the size of the problem. For example, in previous DPP-based methods used to sample diverse subsets~\cite{kwok2012priors,mariet2016diversity}, they need to calculate the values of $\det(G)$ to obtain the sampling probability for each subset, where $m$ usually scales with the number of classes or number of hidden units in networks. However, in our method, we have $G=\tilde{M}_{\backslash y}^{\top} \tilde{M}_{\backslash y}$ and $m=K$, where $K$ grows much slower with the scale of problem~\cite{russakovsky2015imagenet}. This property enables our method to scale to most of the modern machine learning tasks and avoid similar computational obstacle encountered in previous work. As experimentally shown in Section~\ref{Experiments}, our method is also compatible with other defenses acting on individual networks, e.g., adversarial training~\cite{goodfellow6572explaining,madry2017towards}, and thus our method could be an orthogonal approach to further improve robustness for ensemble models. These results verify that ensemble diversity is indeed an effective component to consider when designing stronger defenses.


\vspace{-0.cm}
\subsection{Theoretical Analyses on ADP Training}
\label{Theoretical}
\vspace{-0.1cm}

The minimization problem in the ADP training on a single input-label pair $(x,y)$ can be formally denoted as
\vspace{-0.1cm}
\begin{equation}
    \begin{split}
\min_{\theta} &\text{     }{\Loss}_{\ECE}-\ADP_{\alpha,\beta}\\
\stt &\text{     } 0\leq F_{j}^{k} \leq 1\text{, }\sum_{j\in[L]}F_{j}^{k}=1\text{,}
\end{split}
    \label{equation:3}
\end{equation}
\vspace{-0.5cm}

where $\theta=\{\theta^k\}_{k\in[K]}$ denotes all the trainable parameters in the ensemble. For multiple input-label pairs, the objective function is simply a summation of each single function as in Algorithm~\ref{algo:2}. However, to simplify the analysis, we focus on the optimal solution in the space of output predictions $\mathbb{F}=\{F^k\}_{k\in[K]}$, and assume that the representation capacity of each network is unlimited~\cite{Hornik1989}. Therefore, we ignore the specific mapping from $x$ to each $F^k$, and redefine the above minimization problem with respect to $\mathbb{F}$, rather than $\theta$. Below we will theoretically analyze how the values of the two hyperparameters $\alpha$ and $\beta$ influence the solution for the minimization problem. If $\alpha=0$, the ADP regularizer degenerates to the LED part. In this case, the solution will be one-hot vectors $1_{y}$ just as there is no regularizer, as shown in the following theorem:

\begin{theorem}
(Proof in Appendix A) If $\alpha=0$, then $\forall \beta\geq 0$, the optimal solution of the minimization problem~(\ref{equation:3}) satisfies the equations $F^{k}=1_y$, where $k\in[K]$.
\label{theorem:3}
\vspace{-0.15cm}
\end{theorem}

The conclusion means that the LED part in the ADP regularizer cannot work alone without the ensemble entropy part, indicating that the ensemble entropy part is necessary. In a nutshell, this is because we define the LED part on the \emph{normalized} predictions $\tilde{F}^{k}_{\backslash y}$ corresponding to untrue labels. Thus the LED part can only affect the mutual angles of these untrue predictions. The summation of all unnormalized untrue predictions is $1-F^{k}_{y}$, and in this case, the summation will only be influenced by the ECE loss part, which leads to one-hot optimal solution $1_{y}$. On the other hand, if $\beta=0$, the ADP regularizer degenerates to the ensemble entropy, which results in similar optimal solutions on ensemble prediction $\F_en$ as label smoothing~\cite{szegedy2016rethinking}. The equivalent smoothing factor is $1-\F_en_{y}$, and the conclusions are formally stated in the following theorem:

\begin{theorem}
(Proof in Appendix A) When $\alpha>0$ and $\beta=0$, the optimal solution of the minimization problem~(\ref{equation:3}) satisfies the equations $F_{y}^{k}=\F_en_y$, $\F_en_j=\frac{1-\F_en_y}{L-1}$ and
\vspace{-0.2cm}
\begin{equation}
    \frac{1}{\F_en_y}=\frac{\alpha}{K}\log\frac{{\F_en}_y(L-1)}{1-{\F_en}_y}\text{,}
    \label{equation:2}
\end{equation}
where $k\in[K]$ and $j\in [L]\backslash \{y\}$.
\label{theorem:2}
\end{theorem}

\begin{figure}[t]
\begin{center}
\vspace{-0.2cm}
\includegraphics[width=1.0\columnwidth]{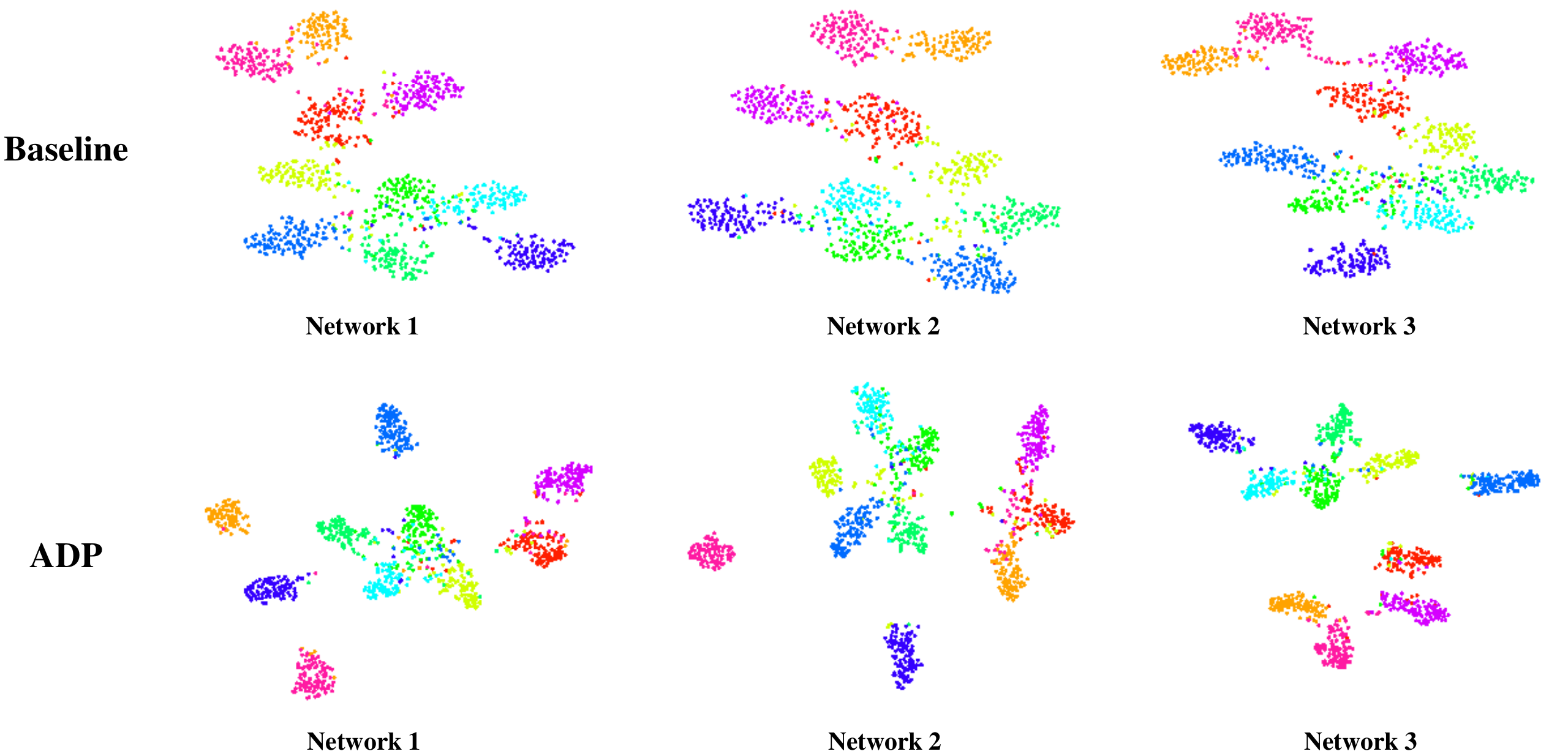}
\vspace{-0.7cm}
\caption{t-SNE visualization results on the final hidden features of each individual network. The input is the test set of CIFAR-10.}
\label{fig:4}
\vspace{-0.2cm}
\end{center}
\end{figure}

Here Eq.~(\ref{equation:2}) can assist us in selecting proper values for the hyperparameter $\alpha$. For example, if we want to train an ensemble of $5$ individual networks ($K=5$) on ImageNet ($L=1000$) with equivalent label smoothing factor $0.1$ ($\F_en_{y}=0.9$), then we can calculate from Eq.~(\ref{equation:2}) that $\alpha\approx 0.61$. The difference of the solutions in Theorem~\ref{theorem:2} from those of label smoothing is that there is an extra constraint for each individual prediction: $F_{y}^{k}=\F_en_y$. This guarantees the consistency of all maximal predictions with the true label. Besides, it is worthy to note that for any $j\in [L]\backslash \{y\}$, the ensemble entropy part does not directly regularize on the non-maximal predictions $F_{j}^{k}$, but regularize on their mean, i.e., the ensemble prediction $\F_en_{j}$. This leaves degrees of freedom on the optimal solutions of $F_{j}^{k}$, and the effect of the LED part is to further regularize on them: if $L-1$ is divisible by $K$, i.e., $K\mid (L-1)$, the non-maximal predictions of each network are mutually orthogonal, as intuitively shown in the right panel of Fig.~\ref{fig:2}, where $K=3$ and $L=10$. We mathematically denote the optimal solution in this case as in the following corollary:

\begin{corollary}
If there is $K\mid (L-1)$, then $\forall\alpha,\beta>0$, the optimal solution of the minimization problem~(\ref{equation:3}) satisfies the Eq.~(\ref{equation:2}). Besides, let $S=\{s_1,,\cdots,s_K\}$ be any partition of the index set $[L]\backslash \{y\}$, where $\forall k\in[K]$, $\left|s_k\right|=\frac{L-1}{K}$. Then the optimal solution further satisfies:
\vspace{-0.1cm}
\begin{eqnarray}
F_j^k=\begin{cases}
\frac{K(1-\F_en_y)}{L-1}\text{,}& j\in s_k\text{,}\\
\F_en_y\text{,}& j=y\text{,}\\
0\text{,}& otherwise\text{.}
\end{cases}
\label{equ:8}
\end{eqnarray}
\vspace{-0.3cm}
\label{corollary:1}
\end{corollary}
The index subset $s_k$ includes all the indexes except for $y$ that the $k$-th classifier predicts non-zero probability. Note that the partition $S$ is adaptive concerning the input, which means that for two inputs $x_1$ and $x_2$ of the same label $y$, the partition $S_1$ and $S_2$ could be different. For example, for dog image \#1 and dog image \#2, an individual network may both assign $0.7$ probability to label dog, and separately assign $0.1$ probability to labels car, frog, ship for image \#1 and plane, truck, cat for image \#2. This is why the ADP regularizer is named as \emph{adaptive}, and this adaptivity can avoid imposing an artificial bias on the relationship among different classes. When $L$ is large, the optimal solution will still approximately satisfy the conclusion if $K\nmid (L-1)$. As shown in Table~\ref{table:9} of our experiments, even when $L$ is not large ($L=10$), the ADP method can still work well in the case of $K\nmid (L-1)$. After training by the ADP method, the non-maximal predictions of each network will tend to be mutually orthogonal, and further leads to different feature distributions and decision domains as shown in Fig.~\ref{fig:4}, where more technical details can be found in Section~\ref{nor_acc}.


\vspace{-0.18cm}
\section{Experiments}
\label{Experiments}
\vspace{-0.1cm}
We now provide the experimental results to demonstrate the effectiveness of the ADP training on improving adversarial robustness under different attack methods, while maintaining state-of-the-art accuracy on normal examples.

\vspace{-0.15cm}
\subsection{Setup}
\vspace{-0.1cm}
We choose three widely studied datasets---MNIST, CIFAR-10 and CIFAR-100~\cite{lecun1998gradient,Krizhevsky2012}. MNIST
consists of grey images of handwritten digits in classes 0 to 9, and CIFAR-10 consists of color images in 10 different classes. CIFAR-100 is similar to CIFAR-10 except that it has 100 classes. Each dataset has 50,000 training images and 10,000 test images. The pixel values of images are scaled to be in the interval $[0,1]$. The normal examples in our experiments refer to all the ones in the training and test sets. The code is available at \href{https://github.com/P2333/Adaptive-Diversity-Promoting}{https://github.com/P2333/Adaptive-Diversity-Promoting}.

\vspace{-0.15cm}
\subsection{Performance on Normal Examples}
\label{nor_acc}
\vspace{-0.1cm}
We first evaluate the performance in the normal setting. On each dataset, we implement the ensemble model consisting of three Resnet-20 networks~\cite{he2016deep}, where $K=3$ satisfies the condition $K\mid(L-1)$ for $L=10$ and $L=100$. In Section~\ref{sec:45}, we will investigate on the case when $K\nmid(L-1)$. Our baseline training method is to train the ensemble with the ECE loss, which is equivalent to ADP training with $\alpha=\beta=0$. As analyzed in Section~\ref{train_strategies}, simultaneously training all networks with the ECE loss is also equivalent to independently training each one with the corresponding CE loss, which is the most commonly applied strategy on training DNN ensembles~\cite{russakovsky2015imagenet,goodfellow2016deep}. About our method, we test two hyperparameter settings: \textbf{$\ADP_{2,0}$} sets $\alpha=2$ and $\beta=0$ corresponding to the case in Theorem~\ref{theorem:2}, where the value of $\alpha$ is chosen according to Eq.~(\ref{equation:2}). This is a reference setting to demonstrate the effect of the ensemble entropy part alone in the ADP regularizer. \textbf{$\ADP_{2,0.5}$} sets $\alpha=2$ and $\beta=0.5$ corresponding to the case in Corollary~\ref{corollary:1}, which is a full version of our method. Later we may omit the subscript as \textbf{$\ADP$} to denote this setting without ambiguity.

\begin{table}[t]
  \vspace{-0.45cm}
  \caption{Classification error rates (\%) on the test set of each dataset. The ensemble model consists of three Resnet-20 networks.}
  \label{table1}
  \begin{center}
  \begin{small}
  \begin{tabular}{c|c|c|c|c}
   \hline
     Dataset &Classifier & Baseline & $\ADP_{2,0}$ & $\ADP_{2,0.5}$ \\
    \hline
    \hline
    \multirow{4}{1.5cm}{\textbf{MNIST}}&Net 1& 0.44  & 0.43 & 0.36 \\
    &Net 2& 0.39  & 0.45 & 0.47 \\
    &Net 3& 0.43  & 0.37 & 0.47 \\
    &Ensemble & 0.32  & 0.32 & \textbf{0.28}     \\
    \hline
    \multirow{4}{1.5cm}{\textbf{CIFAR-10}}&Net 1& 8.30  & 9.01 & 9.34 \\
    &Net 2& 8.52  & 9.15 & 9.34 \\
    &Net 3& 8.67  & 9.14 & 9.92 \\
    &Ensemble & 6.78  & 6.74 & \textbf{6.56}   \\
    \hline
    \multirow{4}{1.6cm}{\textbf{CIFAR-100}}&Net 1& 35.25  & - & 39.04 \\
    &Net 2& 35.91  & - & 40.86 \\
    &Net 3& 36.03  &- & 39.00 \\
    &Ensemble & 30.35  &- & \textbf{29.80}   \\
    \hline
      \end{tabular}
  \end{small}
  \end{center}
    \vspace{-0.1cm}
\end{table}

\begin{table*}[t]
\vspace{-.5cm}
  \caption{Classification accuracy (\%) on adversarial examples. Ensemble models consist of three Resnet-20. For JSMA, the perturbation $\epsilon=0.2$ on MNIST, and $\epsilon=0.1$ on CIFAR-10. For EAD, the factor of $L_1$-norm $\beta=0.01$ on both datasets. }
  \begin{center}
  \begin{small}
  \begin{tabular}{c|c|c|c|c|c|c|c|c}
   \hline
   &\multicolumn{4}{c|}{\textbf{MNIST}}&\multicolumn{4}{c}{\textbf{CIFAR-10}}\\
      Attacks & Para.& Baseline & $\ADP_{2,0}$ & $\ADP_{2,0.5}$ & Para.& Baseline & $\ADP_{2,0}$ & $\ADP_{2,0.5}$ \\
     \hline
    \hline
    \multirow{2}{1.5cm}{FGSM}&$\epsilon=0.1$ &78.3&95.5 &\textbf{96.3} & $\epsilon=0.02$ & 36.5 &57.4 & \textbf{61.7}\\
    &$\epsilon=0.2$ &21.5&50.6 & \textbf{52.8} & $\epsilon=0.04$ & 19.4 &41.9 & \textbf{46.2}\\
    \hline
    \multirow{2}{1.5cm}{BIM} &  $\epsilon=0.1$ &52.3&86.4 & \textbf{88.5}&  $\epsilon=0.01$ & 18.5 & 44.0 & \textbf{46.6}\\
    & $\epsilon=0.15$ &12.2& 69.5&\textbf{73.6}& $\epsilon=0.02$ & 6.1 &28.2 & \textbf{31.0}\\
    \hline
    \multirow{2}{1.5cm}{PGD}& $\epsilon=0.1$ &50.7& 73.4 &\textbf{82.8}& $\epsilon=0.01$ & 23.4 &43.2 & \textbf{48.4}\\
    & $\epsilon=0.15$ &6.3&36.2 &\textbf{41.0}& $\epsilon=0.02$ & 6.6 & 26.8& \textbf{30.4}\\
    \hline
    \multirow{2}{1.5cm}{MIM}& $\epsilon=0.1$ &58.3&89.7 &\textbf{92.0}& $\epsilon=0.01$ & 23.8 & 49.6& \textbf{52.1}\\
    & $\epsilon=0.15$ &16.1&73.3 &\textbf{77.5}&$\epsilon=0.02$ & 7.4 & 32.3 & \textbf{35.9}\\
    \hline
    \multirow{2}{1.5cm}{JSMA}&$\gamma=0.3$&84.0&88.0 & \textbf{95.0}& $\gamma=0.05$& 29.5  & 33.0  & \textbf{43.5}\\
    &$\gamma=0.6$&74.0&85.0&\textbf{91.0}&$\gamma=0.1$& 27.5  & 32.0  & \textbf{37.0}\\
    \hline
    \multirow{3}{1.5cm}{C\&W}& $c=0.1$&91.6& 95.9&\textbf{97.3}& $c=0.001$& 71.3& 76.3 & \textbf{80.6} \\
    &$c=1.0$&30.6&75.0 &\textbf{78.1}&$c=0.01$& 45.2&50.3 & \textbf{54.9} \\
    &$c=10.0$&5.9& 20.2 &\textbf{23.8}&$c=0.1$& 18.8& 19.2& \textbf{25.6} \\
    \hline
        \multirow{2}{1.5cm}{EAD}&$c=5.0$&29.8& 91.3& \textbf{93.4}& $c=1.0$ & 17.5&64.5 & \textbf{67.3}\\
    &$c=10.0$&7.3&87.4 & \textbf{89.5}&$c=5.0$ & 2.4&23.4 & \textbf{29.6}\\
    \hline
      \end{tabular}
  \end{small}
  \end{center}
  \label{table:7}
  \vspace{-1cm}
\end{table*}

In Table~\ref{table1}, we show the test error rates on each dataset. For all the three training settings, we use the same training hyperparameters, e.g., learning rate, training epoch, etc. The weight initialization is different for each network. We apply Adam optimizer~\cite{kingma2014adam} with an initial learning rate of $0.001$. Following similar setting as in~\citet{he2016deep}, we separately run the training process for $40$ epochs on MNIST, $180$ epochs on CIFAR-10 and CIFAR-100 with the mini-batch size of $64$ on a Tesla P100 GPU worker. As Table~\ref{table1} shows, although the ADP training causes higher error rates on individual networks, it leads to lower error rates on the ensembles. This verifies that promoting the ensemble diversity as defined in Eq.~(\ref{equation:12}) is also effective to improve ensemble performance in the normal setting~\cite{krogh1995neural,tsymbal2005diversity}. To investigate the influence of the ADP training on the feature distributions, we use the t-SNE method~\cite{maaten2008visualizing} to visualize the final hidden features of each network in Fig.~\ref{fig:4}. We can see that when trained by the baseline method, the three individual networks have learned similar feature distributions. In contrast, when trained by the ADP method, the learned feature distributions are much more different, which makes it more difficult for adversarial examples to transfer among individual networks as demonstrated in Section~\ref{transfer}.  

Furthermore, to demonstrate that the ADP method only results in a marginal increase in training time, we test the training time per epoch with the mini-batch size of $32$ on CIFAR-10. When $K=5$, it takes $204$s/epoch for the baseline method, and $227$s/epoch for the ADP methods; when $K=20$, it takes $692$s/epoch for baseline, and $744$s/epoch for ADP. Thus the ADP methods only increase around $10\%$ training time compared to baseline even when $K=20$.

\begin{table}[t]
  \caption{Classification accuracy (\%) on adversarial examples. Ensemble models consist of three Resnet-20.}
  \begin{center}
  \begin{small}
  \begin{tabular}{c|c|c|c}
   \hline
   &\multicolumn{3}{c}{\textbf{CIFAR-100}}\\
      Attacks & Para.& Baseline & $\ADP_{2,0.5}$\\
    \hline
    \hline
    \multirow{2}{1.5cm}{BIM} &  $\epsilon=0.005$ &21.6 & \textbf{26.1}\\
    & $\epsilon=0.01$ &10.1&\textbf{14.8}\\
    \hline
    \multirow{2}{1.5cm}{PGD}& $\epsilon=0.005$ &26.6 &\textbf{32.1}\\
    & $\epsilon=0.01$ &11.7 &\textbf{18.3}\\
    \hline
    \multirow{2}{1.5cm}{MIM}& $\epsilon=0.005$ &24.2 &\textbf{29.4}\\
    & $\epsilon=0.01$ &11.2 &\textbf{17.1}\\
    \hline
      \end{tabular}
  \end{small}
  \end{center}
  \label{table:8}
   \vspace{-0.2cm}
\end{table}

\vspace{-0.1cm}
\subsection{Performance under White-box Attacks}
\vspace{-0.1cm}
\label{ensemble_attack}
In the adversarial setting, there are mainly two threat models~\cite{carlini2017adversarial}: \emph{white-box adversaries} know all the information about the classifier models, including training data, model architectures and parameters; \emph{black-box adversaries} know the classification tasks, but have no access to the information about the classifier models.

In this sub-section, we test the performance of ensemble models defending white-box attacks. We apply the attack methods introduced in Section~\ref{attacks}. The results are reported in Table~\ref{table:7} and Table~\ref{table:8}, where we test each attack on two or three representative values of parameters. The iteration steps are set to be $10$ for BIM, PGD, and MIM, with the step size equals to $\epsilon/10$. The iteration steps are set to be $1,000$ for C\&W and EAD, with the learning rate of $0.01$.
We separately test the ensemble models trained by the baseline, $\ADP_{2,0}$ and $\ADP_{2,0.5}$ methods. As expected, the results verify that our method significantly improves adversarial robustness. However, the improvements are more prominent on MNIST and CIFAR-10 than those on CIFAR-100. This can be explained by Eq.~(\ref{equ:8}): the non-maximal predictions scale with the factor $\frac{1}{L-1}$, which could result in a numerical obstacle in the training phase when $L$ is large. A potential way to solve this problem is to use the temperature scaling method~\cite{guo2017calibration}, as discussed in Appendix B.2. An interesting phenomenon is that when the LED part is inactive as in the $\ADP_{2,0}$ setting, the learned ensemble diversity is still much larger than the baseline (detailed in Appendix B.3). This is because the ensemble entropy part expands the feasible space of the optimal solution, while leaving degrees of freedom on the optimal non-maximal predictions $F_{j}^{k}$, as shown in Theorem~\ref{theorem:2}. Then the ensemble diversity is unlikely to be small in this case. However, the existence of the LED part further explicitly encourages the ensemble diversity, which leads to better robustness.

\begin{table}[t]
  \caption{Classification accuracy (\%): \textbf{$\text{AdvT}_\text{FGSM}$} denotes adversarial training (AdvT) on FGSM, \textbf{$\text{AdvT}_\text{PGD}$} denotes AdvT on PGD. $\epsilon=0.04$ for FGSM; $\epsilon=0.02$ for BIM, PGD and MIM.}
  \begin{center}
  \begin{small}
  \begin{tabular}{c|c|c|c|c}
   \hline
   &\multicolumn{4}{c}{\textbf{CIFAR-10}}\\
      Defense Methods & FGSM &BIM & PGD & MIM\\
    \hline
    \hline
    $\text{AdvT}_\text{FGSM}$&39.3 &19.9 &24.2&24.5\\
    $\text{AdvT}_\text{FGSM}+\ADP_{2,0.5}$&\textbf{56.1} & \textbf{25.7}&\textbf{26.7}&\textbf{30.6}\\
    \hline
    $\text{AdvT}_\text{PGD}$&43.2 &27.8 &32.8& 32.7\\
    $\text{AdvT}_\text{PGD}+\ADP_{2,0.5}$&\textbf{52.8} &\textbf{34.0}&\textbf{36.2} &\textbf{38.8}\\
    \hline
      \end{tabular}
  \end{small}
  \end{center}
  \label{table:10}
   \vspace{-0.3cm}
\end{table}

\begin{figure*}[t]
\begin{center}
\vspace{-0.3cm}
\hspace{-0.15cm}
\includegraphics[width=2.1\columnwidth]{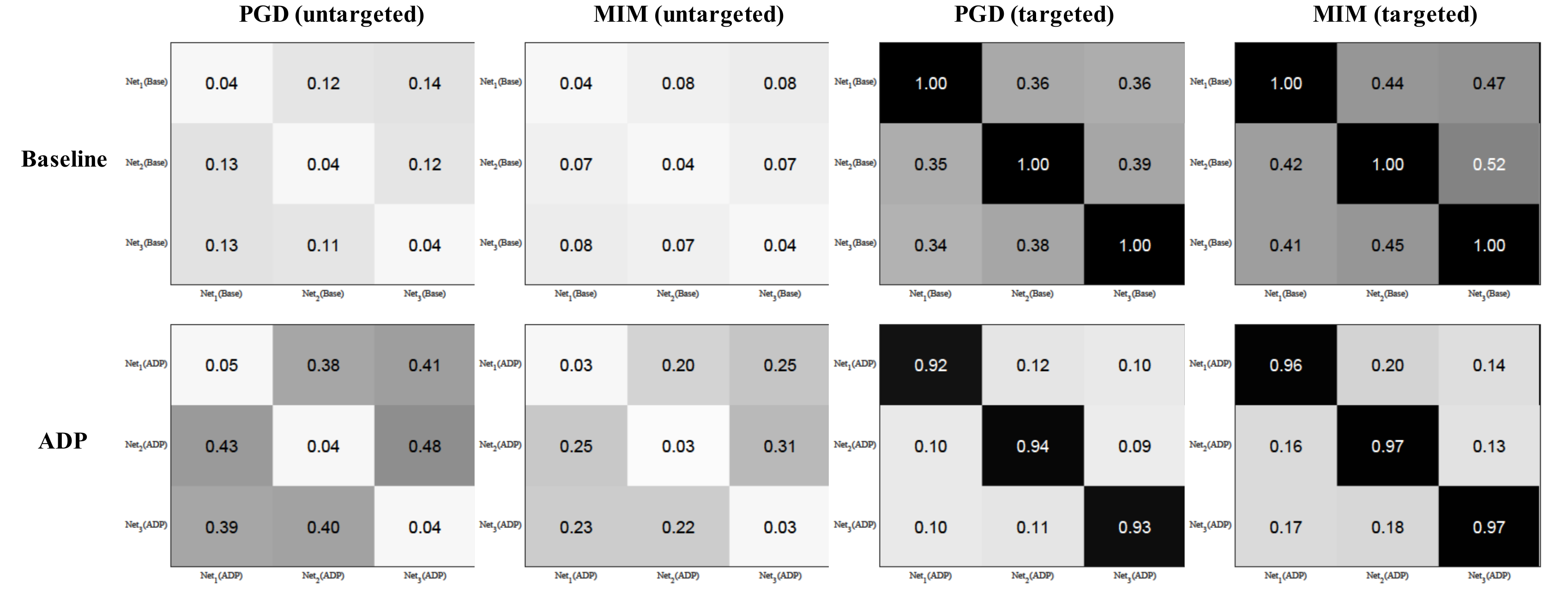}
 \vspace{-0.8cm}
\caption{Adversarial transferability among individual models on CIFAR-10. For untargeted attacks, the values are the classification accuracy; For targeted attacks, those are the success rate of adversarial examples fooling classifiers to predict target classes.}
\label{fig:3}
\end{center}
  \vspace{-0.8cm}
\end{figure*}

To show that our method is an orthogonal approach concerning other defenses acting on a single network, we test the compatibility of our method with adversarial training (AdvT), which is the most widely studied defense method~\cite{kurakin2018competation}. AdvT augments the training data with adversarial examples in each mini-batch, where the two quite common attacks used to craft these adversarial examples are FGSM~\cite{goodfellow6572explaining} and PGD~\cite{madry2017towards}. In Table~\ref{table:10}, we demonstrate the classification accuracy of enhanced ensemble models on CIFAR-10. We uniformly sample the perturbation $\epsilon$ from the interval $[0.01,0.05]$ when performing AdvT as in~\citet{kurakin2016atscale}. The ratio of adversarial examples and normal ones in each mini-batch is 1:1, with the batch size of $128$. The results demonstrate that our method can further improve robustness for ensemble models, with little extra computation and is universal to different attacks. Besides, when comparing the results in Table~\ref{table:7} and Table~\ref{table:10}, we can find that AdvT can also further improve the performance of our method, which means they are indeed complementary.

\begin{figure}[t]
\begin{center}
\vspace{-0.cm}
\includegraphics[width=1.0\columnwidth]{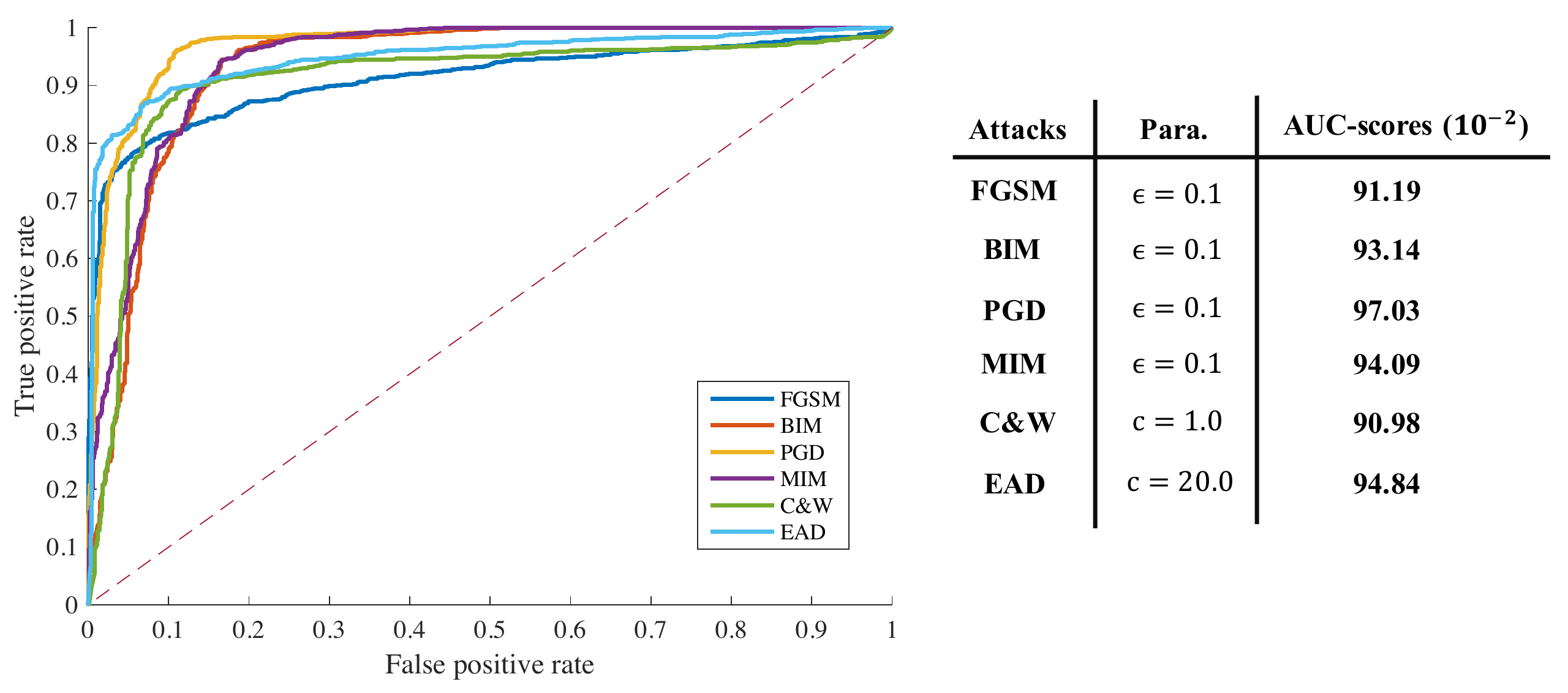}
  \vspace{-0.85cm}
\caption{The ROC-curves on 1,000 test examples and 1,000 adversarial examples of CIFAR-10. The model is trained by $\ADP_{2,0.5}$.}
\label{fig:19}
\end{center}
  \vspace{-.2cm}
\end{figure}

\vspace{-0.cm}
\subsection{Transferability among Individual Models}
\vspace{-0.cm}

\label{transfer}
Due to the transferability of adversarial examples among models~\cite{papernot2016limitations}, black-box adversaries can craft adversarial examples based on substitute models and then feed these examples to original models to perform the attack. In this sub-section, we show that the ADP training can mitigate the transferability of adversarial examples among the members in the ensemble, as demonstrated in Fig.~\ref{fig:3}. For each matrix, the $(i,j)$ element is the result of using $i$-th network as the substitute model to craft adversarial examples and feeding to $j$-th network as the original model. We apply PGD and MIM as the attack methods, which are the two most commonly used attacks in the black-box setting~\cite{kurakin2018competation}. The perturbation parameters for both attacks are set to be $\epsilon=0.05$. For a complete analysis, we test untargeted mode and targeted mode. For untargeted mode, adversaries only aim to make the models predict wrong labels, and the evaluation criterion is the classification accuracy of the models on adversarial examples. For targeted mode, the evaluation criterion becomes the success rate of fooling the models to predict specific target label. As the results indicate, the ADP training can significantly alleviate the transferability among the members of the ensembles in both untargeted and targeted modes, which can further lead to better robustness for the ensemble models.

\vspace{-0.15cm}
\subsection{Ensemble Diversity as the Detection Metric}
\vspace{-0.1cm}
When the adversarial perturbation becomes larger, neither existing defenses nor our method can avoid misclassifying on white-box adversarial examples~\cite{athalye2018obfuscated}. Then another possible solution is detecting and filtering out adversarial examples instead~\cite{pang2017towards}. In Fig.~\ref{fig:19}, we show the ROC-curves and corresponding AUC-scores when applying ensemble diversity as the detection metric to filter out adversarial examples. The models are trained by $\ADP_{2,0.5}$. Here the parameters used for all attacks are much larger than those in Table~\ref{table:7}, which can make trained models misclassify with nearly 100\% rate. The results show the effectiveness of ensemble diversity as the detection metric on ADP-trained models, which generalizes our method to a more extensible defense in the adversarial setting.

\begin{table}[t]
\vspace{0.1cm}
  \begin{center}
  \begin{small}
  \begin{tabular}{c|c|c|c}
   \hline
   &\multicolumn{3}{c}{\textbf{CIFAR-10}}\\
      Attacks & Para.& Baseline & $\ADP_{2,0.5}$\\
    \hline
    \hline
    Normal & - & 93.6 &\textbf{93.8}\\
    FGSM &  $\epsilon=0.02$ &42.0 & \textbf{58.4}\\
    BIM &  $\epsilon=0.01$& 31.6&\textbf{41.8}\\
    PGD& $\epsilon=0.01$ &37.4 &\textbf{44.2}\\
    MIM& $\epsilon=0.01$ &37.1 &\textbf{47.5}\\
    C\&W& $c=0.01$ &52.3 &\textbf{56.5}\\
    EAD& $c=1.0$ &20.4 &\textbf{65.3}\\
    \hline
      \end{tabular}
  \end{small}
  \end{center}
  \vspace{-0.2cm}
  \caption{Classification accuracy (\%) on adversarial examples. Ensemble models consist of \emph{five} Resnet-20 such that $5\nmid 9$.}
  \label{table:9}
    \vspace{-0.05cm}

\end{table}

\vspace{-0.15cm}
\subsection{The Cases of $K\nmid(L-1)$}
\label{sec:45}
\vspace{-0.125cm}
To further investigate the flexibility of our method, we consider the cases of $K\nmid(L-1)$. We test on CIFAR-10 with $L=10$ and construct an ensemble consisting of five individual members, i.e., $K=5$. It is obvious that $5\nmid9$. The results are demonstrated in Table~\ref{table:9}. We can find that even if the condition $K\mid(L-1)$ in Corollary~\ref{corollary:1} does not hold, the ADP training can still largely outperform the baseline method, which suggests that our method is practically insensitive to the relationship of $K$ and $L$, even if $L$ is not large, and thus can be applied in more general cases.

\vspace{-0.2cm}
\section{Conclusion}
\vspace{-0.125cm}
In this paper, we propose the ADP training method. Compared to previous efforts that focus on enhancing a single network, we provide an orthogonal approach to further improve a practical ensemble defense, which implies that ensemble diversity could be an important consideration to promote when designing robust ensemble systems.

\clearpage
\section*{Acknowledgements}
The authors would like to thank Chongxuan Li for helpful comments. This work was supported by the National Key
Research and Development Program of China (No. 2017YFA0700904), NSFC Projects (Nos. 61620106010, 61621136008, 61673241, 61571261), Beijing NSF Project (No. L172037), DITD Program JCKY2017204B064, Tiangong Institute for Intelligent Computing, Beijing Academy of Artificial Intelligence (BAAI), NVIDIA NVAIL Program, and the projects from Siemens and Intel.

\bibliography{reference}
\bibliographystyle{icml2019}

\clearpage
\appendix
\section{Proof}
We give the proofs of the theorems proposed in the paper.
\subsection{Proof of Theorem 1}
When $\alpha=0$ and $\beta\geq0$, the optimization problem can be formalized with constraints as
\[
\begin{split}
\min_{\mathbb{F}} &\text{     }{\Loss}_{\ECE}-\beta\cdot\log\left[\det(\tilde{M}_{\backslash y}^{\top} \tilde{M}_{\backslash y})\right]\\
\stt &\text{     } 0\leq F_{j}^{k} \leq 1\text{,}\\
&\text{     } \sum_{j\in[L]}F_{j}^{k}=1\text{,}
\end{split}
\]
where $k\in[K],\text{ }j\in[L]$. Note that in the objective function, the first term ${\Loss}_{\ECE}$ depends on the predictions on label $y$, i.e., $F^{k}_{y}$. The second term $-\log\left[\det(\tilde{M}_{\backslash y}^{\top} \tilde{M}_{\backslash y})\right]$ depends on the normalized non-maximal predictions $\tilde{F}^{k}_{\backslash y}$. Because $\forall i,j$, $F^{i}_{y}$ and $F^{j}_{y}$ are mutually independent, $F^{i}_{y}$ and $\tilde{F}^{i}_{\backslash y}$ are also mutually independent (since the normalization), the two terms in the objective function can separately achieve their own minimum. Therefore, the optimal solution of the objective function will tend to satisfies the equations $F^k=1_y$, where $k\in[K]$.

\qed

\subsection{Proof of Theorem 2}
When $\alpha>0$ and $\beta=0$, the optimization problem can be formalized with constraints as
\[
\begin{split}
\min_{\mathbb{F}} &\text{     }{\Loss}_{\ECE}-\alpha\cdot\H_en(\F_en)\\
\stt &\text{     } 0\leq F_{j}^{k} \leq 1\text{,}\\
&\text{     } \sum_{j\in[L]}F_{j}^{k}=1\text{,}
\end{split}
\]
where $k\in[K],\text{ }j\in[L]$.
Then the Lagrangian is
\begin{multline*}
    L={\Loss}_{\ECE}-\alpha\cdot\H_en(\F_en)+\sum_{k\in[K]}\omega_{k}(1-\sum_{j\in[L]}F_{j}^{k})\\+\sum_{k\in[K]}\sum_{j\in[L]}\left[\beta_{k,j}F_{j}^{k}+\gamma_{k,j}(1-F_{j}^{k})\right]\text{,}
\end{multline*}
where $\beta_{k,j}\leq0$, $\gamma_{k,j}\leq0$. The partial derivatives for $F_{j}^{k}$ are
\[
\begin{split}
    &\frac{\partial L}{\partial F_{y}^{k}}=-\frac{1}{F_{y}^{k}}+\frac{\alpha}{K}\left[1+\log\F_en_{y}\right]-\omega_k+\beta_{k,y}-\gamma_{k,y}\text{,}\\
    &\frac{\partial L}{\partial F_{j}^{k}}=\frac{\alpha}{K}\left[1+\log\F_en_{j}\right]-\omega_k+\beta_{k,j}-\gamma_{k,j}\text{, }\forall j\neq y\text{.}
\end{split}
\]
According to the KKT conditions for the optimal solution, we have $\forall k\in[K]\text{, }j\in[L]$,
\[
\begin{split}
&\frac{\partial L}{\partial F_{j}^{k}}=0\text{,}\\
&\beta_{k,j}F_{j}^{k}=0\text{,}\\
&\gamma_{k,j}(1-F_{j}^{k})=0\text{.}
\end{split}
\]
Consider the optimal solutions in $(0,1)^{L\times K}$, then all $\beta_{k,j}$ and $\gamma_{k,j}$ equal to zero. Now we have
\[
\begin{split}
&-\frac{1}{F_{y}^{k}}+\frac{\alpha}{K}\left[1+\log\F_en_{y}\right]=\omega_k\text{,}\\
&\frac{\alpha}{K}\left[1+\log\F_en_{j}\right]=\omega_k\text{, }\forall j\neq y\text{,}
\end{split}
\]
and from the second equations, we can derive $\forall j\neq y$,
\[
\begin{split}
    \F_en_j=\exp{(\frac{\omega_k K}{\alpha}-1)}&\Longrightarrow\sum_{j\neq y}\F_en_j=\sum_{j\neq y}\exp{(\frac{\omega_k K}{\alpha}-1)}\\
    &\Longrightarrow 1-\F_en_{y}=(L-1)\exp{(\frac{\omega_k K}{\alpha}-1)}\\
    &\Longrightarrow \omega_{k}=\frac{\alpha}{K}\left[1+\log(\frac{1-\F_en_{y}}{L-1})\right]\text{,}
\end{split}
\]
and this also shows that $\forall i,j \neq y$, there is $\F_en_i=\F_en_j=\frac{1-\F_en_y}{L-1}$. There further is $\forall k\in [K]$,
\[
\frac{1}{F_{y}^{k}}=\frac{\alpha}{K}\log\left[\frac{\F_en_{y}(L-1)}{1-\F_en_{y}}\right]\text{.}
\]
Thus for $\forall k,l\in [K]$, $F_y^{k}=F_y^{l}=\F_en_y$, and finally
\[
\frac{1}{\F_en_y}=\frac{\alpha}{K}\log\left[\frac{\F_en_{y}(L-1)}{1-\F_en_{y}}\right]\text{.}
\]

\qed

\subsection{Proof of Corollary 1}
It is easy to see that the negative LED part $-\log(\ED)$ achieves its minimum if and only if the non-maximal predictions of each individual network are mutually orthogonal. According to the conclusion in Theorem 2, if there is $K\mid(L-1)$, the optimal solution in Corollary 1 can simultaneously make the two terms of the ADP regularizer achieve their own minimum.

\qed

\section{More Analyses}
In this section, we provide more details on the theoretical and practical analyses mentioned in the paper.
\subsection{JS-divergence as the Diversity}
JS-divergence among individual predictions is a potentially plausible definition of ensemble diversity for DNNs. Specifically, we consider in the output space of classifiers, where $F^{k}$ represents a vector variable in $\R^L$. The JS-divergence of $K$ elements in $\mathbb{F}$ is defined as
\begin{equation}
\label{equ:jsd}
    \JSD(\mathbb{F})=\H_en(\F_en)-\frac{1}{K}\sum_{k\in[K]}\H_en(F^{k})\text{.}
\end{equation}
To encourage high values of JS-divergence , we can add a regularization term of it in the objective function. However, when minimizing the objective function, there is neither a closed form solution nor an intuitively reasonable solution, as formally stated in the following theorem:

\begin{theoremappendix}
Given $\lambda>0$, $(x,y)$ be an input-label pair. The minimization problem is defined as
\begin{equation}
    \min_{\mathbb{F}} {\Loss}_{\ECE}-\lambda\cdot\JSD\text{.}
    \label{equation:7}
\end{equation}
Then the problem has no solution in $(0,1)^{L\times K}$.
\label{theorem:1}
\end{theoremappendix}

\textbf{Proof.} The optimization problem can be formalized with constraints as
\[
\begin{split}
\min_{\mathbb{F}} &\text{     }{\Loss}_{\ECE}-\lambda\cdot\JSD(\mathbb{F})\\
\stt &\text{     } 0\leq F_{j}^{k} \leq 1\text{,}\\
&\text{     } \sum_{j\in[L]}F_{j}^{k}=1\text{,}
\end{split}
\]
where $k\in[K],\text{ }j\in[L]$.
Then the Lagrangian is
\begin{multline*}
    L={\Loss}_{\ECE}-\lambda\cdot\JSD(\mathbb{F})+\sum_{k\in[K]}\omega_{k}(1-\sum_{j\in[L]}F_{j}^{k})\\+\sum_{k\in[K]}\sum_{j\in[L]}\left[\beta_{k,j}F_{j}^{k}+\gamma_{k,j}(1-F_{j}^{k})\right]\text{,}
\end{multline*}
where $\beta_{k,j}\leq0$, $\gamma_{k,j}\leq0$. The partial derivatives for $F_{j}^{k}$ are
\[
\begin{split}
&\frac{\partial L}{\partial F_{y}^{k}}=-\frac{1}{F_{y}^{k}}+\frac{\lambda}{K}\left[\log\F_en_{y}-\log F_{y}^{k}\right]-\omega_k+\beta_{k,y}-\gamma_{k,y}\text{,}\\
&\frac{\partial L}{\partial F_{j}^{k}}=\frac{\lambda}{K}\left[\log\F_en_{j}-\log F_{j}^{k}\right]-\omega_k+\beta_{k,j}-\gamma_{k,j}\text{, }\forall j\neq y\text{.}
\end{split}
\]
According to the KKT conditions for the optimal solution, similar to the proof of Theorem 2, we can derive $\forall j\neq y$,
\[
\begin{split}
    \F_en_j=F_{j}^{k}\exp{(\frac{\omega_k K}{\lambda})}&\Longrightarrow\sum_{j\neq y}\F_en_j=\sum_{j\neq y}F_{j}^{k}\exp{(\frac{\omega_k K}{\lambda})}\\
    &\Longrightarrow 1-\F_en_{y}=(1-F_{y}^{k})\exp{(\frac{\omega_k K}{\lambda})}\\
    &\Longrightarrow \omega_{k}=\frac{\lambda}{K}\log(\frac{1-\F_en_{y}}{1-F_{y}^{k}})\text{,}
\end{split}
\]
further combine with the first equation, there is
\begin{equation}
    \frac{\lambda}{K}\log\left[\frac{\F_en_{y}(1-F_{y}^{k})}{F_{y}^{k}(1-\F_en_{y})}\right]=\frac{1}{F_{y}^{k}}\text{.}
\label{appendix_equ:1}
\end{equation}
Since $\F_en_{y}=\frac{1}{K}\sum_{k\in[K]}F_{y}^{k}$, Eq.~(\ref{appendix_equ:1}) cannot holds for all $k\in[K]$, there is no optimal solution in $(0,1)^{L\times K}$.

\qed

Therefore, it is difficult to appropriately select $\lambda$ and a balance between accuracy and diversity, which makes it unsuitable to directly define the ensemble diversity as JS-divergence. Note that this dilemma is mainly caused by the second term in the definition of JS-divergence (Eq.~(\ref{equ:jsd})) since it can override the ECE term and lead to the wrong prediction with a low value of the total loss.

\subsection{Temperature Scaling}
\citet{guo2017calibration} propose the temperature scaling (TS) method to calibrate the predictions of neural networks. The TS method simply use a temperature $T>0$ on the logits, and return the predictions as
\[
F=\softmax(\frac{z}{T})\text{,}
\]
where $\softmax(\cdot)$ is the softmax function, $z$ is the logits. Usually the temperature $T$ is set to be $1$ in the training phase, and be larger than $1$ in the test phase~\cite{hinton2015distilling}. To solve the numerical obstacle in the ADP training procedure when $L$ is large, we propose to apply the TS method in an opposite way. Namely, in the training phase, we apply a high value of $T$ to increase the values of non-maximal predictions, then in the test phase we apply $T=1$ to give final predictions. Note that the non-maximal predictions in Corollary 1 equal to $\frac{K(1-\F_en_y)}{L-1}$. Using the TS method is equivalent to reduce $\F_en_y$ in the training phase. Other possible ways to solve the numerical obstacle can be increasing the number of members in the ensemble (increasing $K$) or performing a dropout sampling in the ensemble entropy term of the ADP regularizer (decreasing $L-1$). Further investigation of these solutions is one of our future work.

\subsection{Histogram of the Ensemble Diversity}
\begin{figure}[t]
\begin{center}
\includegraphics[width=1.0\columnwidth]{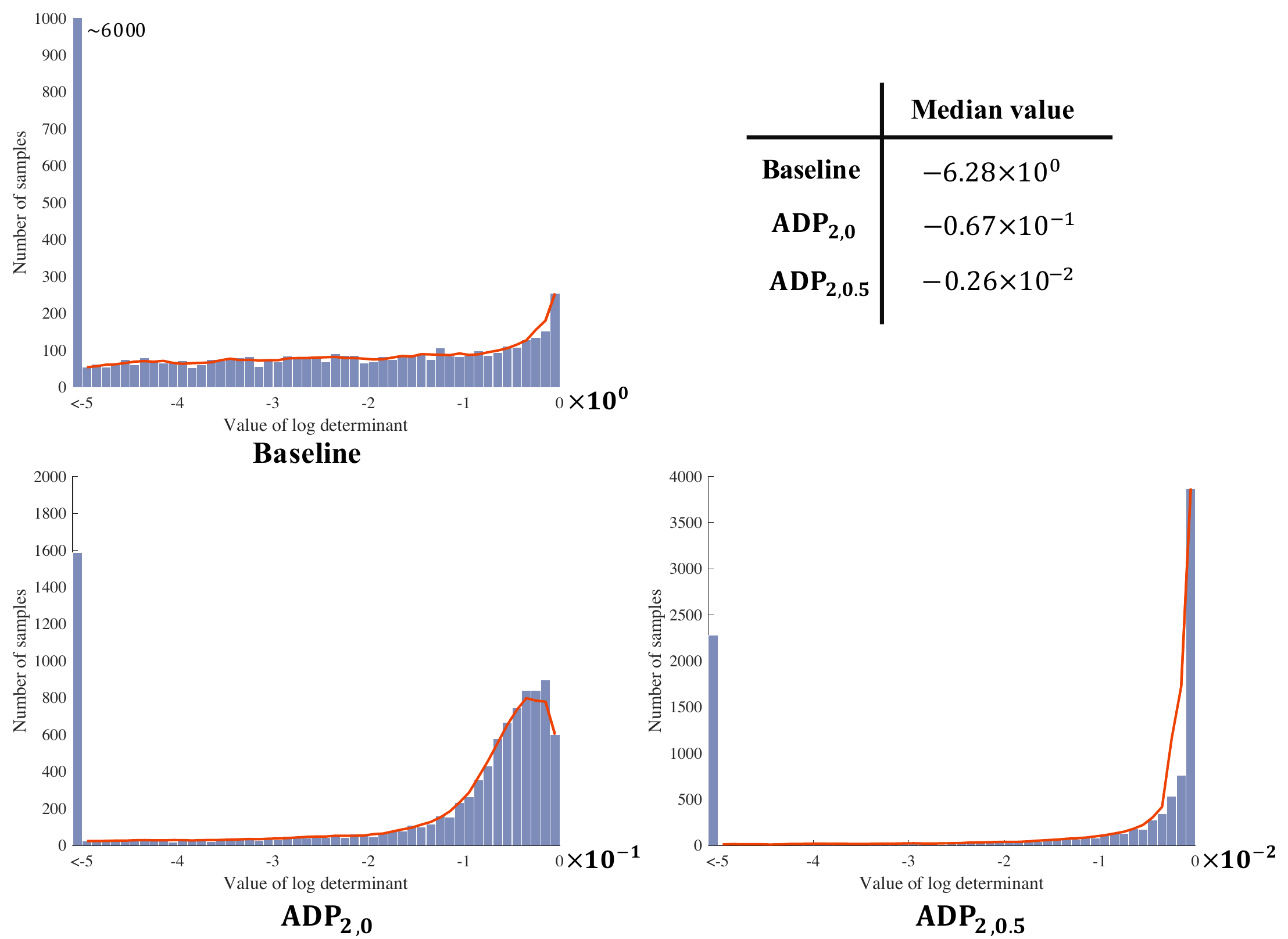}
  \vspace{-0.6cm}
\caption{The histogram of the ensemble diversity values on the test set of CIFAR-10. There are totally 10,000 samples. The tiny table shows the median values of the ensemble diversity.}
\label{fig:7}
\end{center}
  \vspace{0.4cm}
\end{figure}

To further investigate the relationship between ensemble diversity and robustness, we plot the histogram of the logarithm values of ensemble diversity on the test set of CIFAR-10 in Fig.~\ref{fig:7}. The median values of different training methods are shown in the top-right panel. An interesting phenomenon is that when the LED part is inactive as in the $\ADP_{2,0}$ setting, the learned ensemble diversity is still much larger than the baseline. This is because even though the ensemble entropy part of ADP regularizer does not explicitly encourage ensemble diversity, it does expand the feasible space of the optimal solution. Due to the degrees of freedom on the optimal individual predictions $F_{j}^{k}$ in the feasible space, the ensemble diversity is unlikely to be small. However, the existence of the LED part further explicitly encourage the ensemble diversity, as shown in Fig.~\ref{fig:7}.


\end{document}